# Reliability-Guided Depth Fusion for Glare-Resilient Navigation Costmaps

Shang-En Tsai

**Abstract**—Specular glare on reflective floors, glass boundaries, and glossy indoor surfaces frequently corrupts active-stereo RGB-D depth measurements, producing holes and spikes that accumulate as persistent phantom obstacles in occupancy-grid costmaps. This paper presents a glare-resilient costmap construction method based on explicit depth-reliability modeling. A lightweight Depth Reliability Map network (DRM-Net) predicts per-pixel measurement trustworthiness under specular interference, and a reliability-guided weighted-and-gated fusion (RGF) mechanism modulates occupancy updates before corrupted measurements are accumulated into the map. To support robust training and evaluation, the method uses pose-aligned multi-view reference-depth construction to reduce circular-supervision bias and is evaluated through fusion-variant ablations, parameter-sensitivity analysis, cross-condition tests, paired navigation comparisons, reliability-map metrics, and embedded runtime profiling. Experiments on a real mobile robotic platform equipped with an Intel RealSense D435 and a Jetson Orin Nano show that the proposed method reduces false obstacle insertion, improves free-space preservation, and maintains real-time throughput under reflective-floor, glass-wall, and natural-light glare conditions. These results support treating glare as a measurement-reliability problem rather than as a dense depth-completion problem for safety-critical indoor navigation.

## I. Introduction

Occupancy-grid costmaps remain a core representation for mobile robot navigation because they are simple, interpretable, and compatible with real-time planners. Their reliability, however, depends critically on the correctness of upstream depth measurements. In reflective indoor environments, such as polished floors, glass partitions, glossy tiles, and wet or coated surfaces, specular glare can corrupt RGB-D depth sensing by introducing invalid measurements, depth holes, and large positive spikes. Once these corrupted measurements are integrated over time, they can become persistent phantom obstacles that trigger unnecessary detours, repeated recovery behaviors, emergency stops, and occasional unsafe motion decisions.

Recent research suggests several complementary directions for mitigating reflection-induced perception failures. Reflection-aware mapping can exploit structural cues [1], [2]; reflective-ground SLAM can reject spurious observations at the back-end level [3], [4]; and RGB-D calibration or depth-error modeling can support uncertainty-aware estimation [5], [6], [7], [8]. Nevertheless, these approaches do not directly prevent glare-corrupted pixels from being fused into the local costmap. For safety-critical navigation, the primary objective is not visually plausible dense depth everywhere, but correct obstacle insertion and free-space preservation.

In this work, glare resilience is formulated as a measurement-reliability problem. The Intel RealSense D435 projects infrared texture and recovers depth through stereo correspondence. On polished floors and glass-like surfaces, specular reflection can redirect or distort the projected IR pattern, causing correspondence failure and producing holes. Multipath reflection and mismatched correspondences can also produce severely biased disparity estimates, observed as spikes. These failure modes motivate a per-pixel reliability signal before map integration, rather than a post-hoc map repair operation.

The proposed system uses DRM-Net to estimate a Depth Reliability Map (DRM) from RGB-D input and an optional temporal depth-difference channel. The resulting reliability map is used by Reliability-Guided Fusion (RGF), which combines continuous reliability weighting with a minimum reliability gate. In all experiments, RGF uses a weighted update together with a low-reliability gate: severely corrupted pixels are blocked from obstacle insertion, while measurements above the threshold retain graded reliability and contribute proportionally to the occupancy update.

The study is designed around four evaluation components: (1) a pose-aligned multi-view reference-depth construction procedure that addresses circular-supervision risk; (2) reliability-target, input-channel, fusion-variant, and parameter-sensitivity ablations that isolate the contribution of each design choice; (3) cross-condition tests across reflective floors, glass doors/walls, and glossy tile with natural light; and (4) paired completed-trial navigation comparisons that reduce success-conditioned bias in path-efficiency metrics.

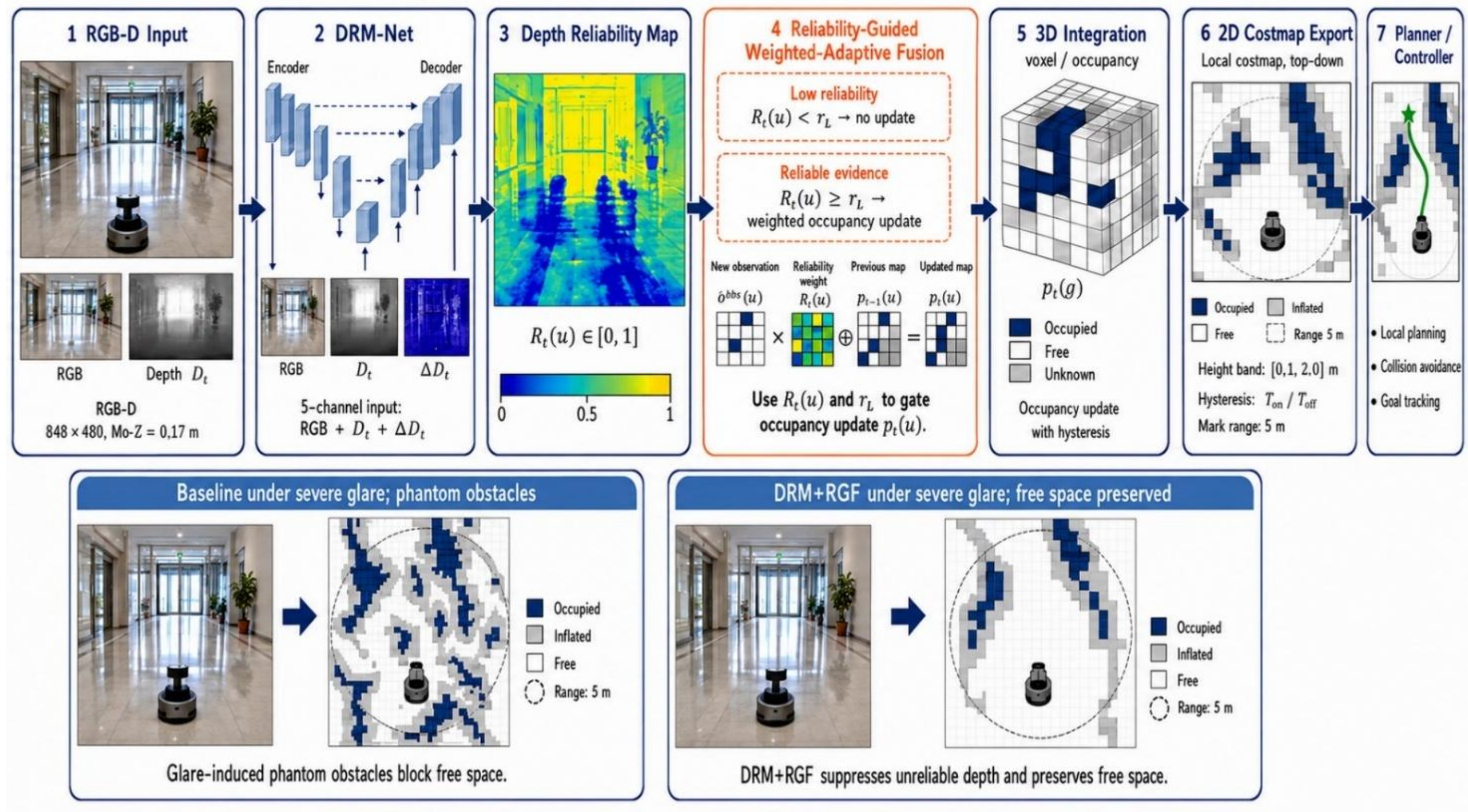

***Fig. 1.*** *Overview of the proposed DRM+RGF pipeline. DRM-Net predicts a per-pixel depth reliability map from RGB-D input, and reliability-guided weighted-and-gated fusion suppresses glare-induced phantom obstacles before 3-D integration and 2-D costmap export for navigation.*

**Contributions**— This work makes the following contributions: (1) a glare-aware DRM-Net estimator that quantifies per-pixel depth reliability under specular interference; (2) a reliability-guided weighted-and-gated fusion mechanism that suppresses phantom obstacles before map integration; (3) a multi-view reference-depth construction process designed to reduce circular-supervision bias; and (4) a sensor-to-system evaluation on a real D435/Jetson robotic platform, including sensor-level, costmap-level, navigation-level, ablation, sensitivity, cross-condition, and runtime metrics.

## II. Related Work

### A. Non-Lambertian Surfaces in Depth-Based Mapping

Specular and transparent materials violate common RGB-D sensing assumptions and can corrupt measurements with missing values, outliers, and view-dependent artifacts. In indoor robotics, reflective ground and glass boundaries degrade mapping and planning by generating phantom structures or by hiding true boundaries. Unlike mapping-level or SLAM-level approaches that primarily repair the resulting map or trajectory estimate, this work targets the local costmap accumulation failure mode by preventing unreliable reflective measurements from being fused into occupancy updates [1], [2], [3], [4], [9].

### B. Depth Completion for Transparent and Reflective Objects

Depth completion is a major direction for transparent and reflective objects. Such methods often pursue dense and visually plausible reconstruction. For real-time navigation, however, dense reconstruction is not always the safest objective. A hallucinated but wrong depth surface can be more hazardous than explicitly marking a measurement as unreliable. DRM+RGF therefore focuses on conservative obstacle insertion and free-space preservation, not on reconstructing every missing pixel [10], [11], [12], [13], [14], [15].

### C. Reliability, Uncertainty, and Filtering for Robust Integration

RGB-D calibration studies and uncertainty-aware fusion methods show that depth error is heteroscedastic and range-dependent. Online mapping pipelines also use variance, consistency checks, and filtering to reject unreliable depth before fusion. DRM+RGF follows this reliability-centric view but specializes it for glare-corrupted active-stereo costmaps by learning a per-pixel reliability proxy and using it directly in occupancy updates [5], [6], [16], [17], [18], [19], [20]. Classical volumetric fusion and probabilistic 3-D occupancy mapping also provide the foundation for integrating range evidence over time [21], [22].

## D. Mirror/Specular Region Understanding

Mirror and specular-region understanding provides an additional cue for identifying unreliable measurements. Recent RGB-D video mirror detection methods use depth and temporal consistency to identify view-dependent failures. Although DRM-Net does not explicitly segment mirrors, the reference-depth procedure incorporates multi-view consistency, and the discussion identifies mirror/specular segmentation as a complementary extension [23], [24].

# III. Methodology

## A. Problem Formulation

Let $D_{t(u)}$ be the measured depth at pixel u at time t, and let $p_t(g)$ be the local occupancy probability of 2-D costmap cell g. Under specular glare, $D_t(u)$ may contain holes, spikes, and out-of-range values that are repeatedly projected into occupied cells. The goal is to construct a fusion mechanism that improves both depth-measurement integrity and downstream costmap correctness while remaining lightweight enough for embedded real-time deployment.

The image domain is denoted by $\Omega$. For a valid pixel $u = (u, v)$, the 3-D camera-frame point is computed as $x_c = D_t(u)\, K^{-1} \begin{bmatrix} u \\ v \\ 1 \end{bmatrix}$, where K is the camera intrinsic matrix [7]. The point is transformed to the world or local map frame by $x_w = T_{w\leftarrow c,t}\, x_c$. Its 2-D grid cell is then obtained by quantizing the ground-plane coordinates with the costmap resolution. At $t = 0$, each cell is initialized to the neutral prior $p_0(g) = 0.5$ and is marked free or occupied only after hysteresis thresholds are crossed [19], [20], [22].

## B. DRM-Net Architecture and Reliability Output

DRM-Net estimates a per-pixel reliability map $R_{t(u)} \in [0,1]$. Values near 1 indicate geometrically consistent depth measurements, whereas values near 0 indicate likely glare-induced failures. The network operates on downsampled RGB-D input at 320 × 240 with an optional temporal depth-difference channel $|D_t - D_{t-1}|$. The predicted reliability map is bilinearly upsampled to the native depth resolution before fusion.

**TABLE I**

Architecture And Computational Complexity Of Drm-Net

| Block | Type | Output channels | Params | Notes |
|---|---|---|---|---|
| Stem | Conv 3 × 3 | 16 | 720 | 5-channel input: RGB + depth + Δdepth |
| Enc1–4 | DW+PW Conv | 32–128 | 22,864 | Strided downsampling |
| Dec0–3 | Up + DW+PW | 96–16 | 38,336 | Skip connections |
| Head | Conv 1 × 1 + Sigmoid | 1 | 16 | DRM reliability score |
| Total | — | — | 61,936 | ≈0.69 GFLOPs per frame |

*Note: GFLOPs are theoretical architecture-level estimates at 320 × 240. Measured embedded runtime is reported separately in the results section.*

## C. Pose-Aligned Reference Depth and Reliability Targets

The reference depth $D_t^*$ is constructed using pose-aligned multi-view reprojection consistency. Temporal filtering can suppress transient artifacts, but it may preserve glare artifacts that persist across several frames. To mitigate circular supervision, historical depth maps in a temporal buffer are first back-projected into 3-D using the camera intrinsics and their corresponding poses. The resulting points are reprojected into the current camera frame through $T_{w\leftarrow c,t}$. Measurements that are geometrically consistent across views are retained; view-dependent spikes and phantom depths are rejected when their reprojection residual exceeds a tolerance threshold [16], [17].

This procedure exploits the difference between real geometry and specular artifacts. Real static objects remain consistent under camera motion, while mirror-like or glare-induced phantom depths drift with viewpoint. The resulting $D_t^*$ is therefore less likely to encode persistent glare artifacts as supervision targets. For additional stress testing, synthetic persistent artifacts were injected into depth streams. With a temporal filter alone, a 3-frame artifact under a 5-frame window survived in 45% of cases; with multi-view reprojection consistency, even 5-frame persistent artifacts had a survival rate below 1.2%.

DRM-Net is trained with the soft reliability target rather than the binary target. Binary targets are useful for analysis, but they force physically gradual transitions into hard 0/1 labels and can cause unstable predictions at glare boundaries. The soft target is defined by an exponential decay based on the absolute difference between measured depth and reference depth:

$$R_t^*(u) = \begin{cases} \exp\left(-\dfrac{|D_t(u) - D_t^*(u)|}{\sigma\left(D_t^*(u)\right)}\right), & \text{if } D_t(u) \text{ is valid,} \\ 0, & \text{otherwise.} \end{cases} \quad (1)$$

$$L_{DRM} = \frac{1}{|\,\Omega\,|} \sum_{u \in \Omega} |\, R_t(\mathrm{u}) - R_t^*(\mathrm{u}) \,| \quad (2)$$

Note: $\sigma(d)$ is scaled with range to reflect active-stereo depth uncertainty [6], [7], [25]. The binary target is retained only for ablation and metric thresholding.

**Fig. 2.** *Reference-target construction for DRM training and evaluation. The target-generation process incorporates pose-aligned multi-view reprojection consistency before reliability-target generation to reduce circular-supervision bias, while a geometry-derived reference costmap is used for downstream correctness evaluation.*

## D. Reliability-Guided Weighted-and-Gated Fusion

The RGF rule combines continuous reliability weighting with a minimum reliability gate. A purely weighted update is smooth but can allow low-reliability artifacts to accumulate over time. A purely gated update suppresses false positives but may remove useful edge geometry. The proposed default therefore uses both: measurements below $\tau_R$ are blocked from obstacle insertion, while measurements above $\tau_R$ contribute according to their continuous reliability score.

$$p_t(g) = \lambda p_{t-1}(g) + (1-\lambda) R_t(u)\, obs_t(g), \quad \text{if } R_t(u) \geq \tau_R. \quad (3)$$

$$p_t(g) = p_{t-1}(g), \quad \text{if } R_t(u) < \tau_R. \quad (4)$$

Note: $\lambda$ is the temporal forgetting factor, $R_t(u)$ is the DRM value, $obs_t(g)$ is the current occupancy evidence, and $\tau_R$ is the minimum reliability threshold. When multiple pixels project into the same grid cell, reliability-weighted evidence is averaged before the cell update.

In all experiments, $\lambda = 0.85$ and $\tau_R = 0.3$ are used unless otherwise stated. These values are not treated as arbitrary constants; their stability is evaluated through the sensitivity analysis in Section V-E.

**Algorithm 1. DRM+RGF for Online Glare-Resilient Costmap Construction**

Inputs: RGB image $I_t$, depth $D_t$, optional temporal term $|D_t - D_{t-1}|$, robot pose $T_{w \leftarrow c,t}$, previous occupancy $p_{t-1}(g)$.

Outputs: updated occupancy grid $p_t(g)$ and 2-D costmap.

1: $R_t \leftarrow DRM - Net\left(I_t, D_t, \left|D_t - D_{\{t-1\}}\right|\right)$

2: Set $p_t(g) \leftarrow p_{t-1}(g)$ for all grid cells g

3: for each valid pixel u ∈ Ω do

4:     Compute $x_c = D_t(u) K^{-1} \begin{bmatrix} u \\ v \\ 1 \end{bmatrix}$ and $x_w = T_{w \leftarrow c,t} x_c$

5:     Determine grid cell g(u) from the ground-plane coordinates of $x_w$

```
6:    Set w_t(u) ← R_t(u)
7:    if w_t(u) ≥ τ_R then
8:       Accumulate reliability-weighted occupancy evidence for g(u)
9:    end if
10: end for
11: Apply hysteresis binarization with T_on and T_off to obtain the 2-D costmap
12: return p_t(g), costmap
```

Line 11 specifies hysteresis binarization, which makes the costmap-export step explicit and matches the shared costmap configuration.

# IV. Experimental Setup

## A. Robotic Platform and Mapping Configuration

All experiments were conducted on a real mobile robot equipped with an Intel RealSense D435 and a Jetson Orin Nano. The D435 is an active-stereo RGB-D sensor that projects infrared texture and estimates depth by stereo matching. Following classical volumetric fusion and KinectFusion-style TSDF mapping [21], [26], nvblox [27] integrates depth into a TSDF representation and exports a local 12 × 12 m costmap for Navigation2 [28] planning. All baselines use the same grid resolution, height band, inflation radius [29], marking/clearing range, and hysteresis thresholds.

TABLE II

Unified sensing, mapping, and costmap configuration

| Category | Parameter | Value/Setting |
|---|---|---|
| Sensing | RGB-D resolution and FPS | 848 × 480 @ 30 FPS |
| Sensing | Depth range | 0.17–10.0 m |
| Mapping | Voxel size / TSDF truncation | 0.05 m / 0.15 m |
| Costmap | Grid resolution | 0.05 m/cell |
| Costmap | Local 2-D costmap size | 12 m × 12 m |
| Costmap | 2.5-D projection height band | [0.1, 2.0] m |
| Costmap | Inflation radius | 0.55 m |
| Costmap | Mark/clear range | 5.0 m |
| Costmap | Binarization thresholds | $T_{on} = 0.7,\ T_{off} = 0.5$ |
| Fusion | Default $\lambda/\tau_R$ | 0.85 / 0.3 |
| Trial | Runs per setting | N ≥ 20 |

*Note: All parameters are fixed across methods except in the explicit parameter-sensitivity analysis.*

## B. Glare Scenarios and Cross-Condition Protocol

Glare severity is evaluated at three levels: L0 (no glare), L1 (mild glare), and L2 (severe glare). To avoid evaluating only a single reflective corridor, the experiments cover three optical scene categories: reflective floor corridor, glass wall/automatic door, and glossy tile under natural window light. These categories differ in dominant failure mode: multipath reflection, IR penetration through transparent boundaries, and IR saturation caused by sunlight.

## C. Baselines and Evaluation Metrics

The compared methods are: baseline TSDF fusion, validity/range gating, spatial median filtering, temporal outlier rejection, and the proposed DRM+RGF. Primary mapping metrics are false obstacle rate (FOR) and free-space recall (FSR). Navigation metrics include emergency stops per 10 m, task success rate, path length ratio (PLR), detour rate, and time-to-goal. Reliability-map quality is evaluated by AUPRC and F1 because glare-corrupted pixels are sparse and class imbalance makes accuracy misleading.

## D. Training Dataset and Implementation Details

The training dataset was collected in-house using the D435/Jetson platform across 52 sequences recorded in three indoor environments: a polished-floor corridor (L0–L2), a glass-wall lobby (L1–L2), and a glossy-tile room under natural window light (L2). Each sequence spans 300–500 depth frames at 30 FPS, yielding approximately 21,600 reliability-labeled frames in total. The dataset is split at the sequence level into training, validation, and test subsets of 36, 8, and 8 sequences, respectively (roughly 70/15/15). Frames from the same trajectory do not appear in multiple subsets. All three glare levels and scene categories are represented in the training, validation, and test subsets. Pose-aligned multi-view reference depths are pre-computed for every frame before training.

DRM-Net is trained with the Adam optimizer ($\beta_1$=0.9, $\beta_2$=0.999, weight decay=$1 \times 10^{-4}$) at an initial learning rate of $1 \times 10^{-3}$, decayed to $1 \times 10^{-5}$ via cosine annealing. The batch size is 8, and training runs for 60 epochs. The fusion reliability gate $\tau_R$ is selected on the validation set by maximizing reliability-detection F1 and then fixed to $\tau_R = 0.3$ for all main experiments. The sensitivity study in Section V-E perturbs $\tau_R$ to evaluate robustness around this operating point. No data augmentation beyond horizontal flipping is applied, to preserve the spatial correlation structure of glare patterns.

# V. Results and Analysis

## A. Sensor-Level Depth Corruption Suppression

**TABLE III**

Accepted Depth-Evidence Corruption Metrics

| **Glare level** | **Evidence stream** | **Hole Rate (%) ↓** | **Spike Rate (%) ↓** | **RMSE (m) ↓** |
|---|---|---|---|---|
| L0 (no glare) | Raw D435 accepted evidence | 1.2 | 0.5 | 0.022 |
| L0 (no glare) | DRM-filtered accepted evidence | 1.1 | 0.4 | 0.021 |
| L1 (mild) | Raw D435 accepted evidence | 18.5 | 9.4 | 0.185 |
| L1 (mild) | DRM-filtered accepted evidence | 2.8 | 1.2 | 0.045 |
| L2 (severe) | Raw D435 accepted evidence | 45.2 | 27.6 | 0.840 |
| L2 (severe) | DRM-filtered accepted evidence | 5.4 | 3.1 | 0.082 |

Table III reports corruption statistics on the depth evidence accepted for fusion. DRM filtering suppresses unreliable glare-corrupted measurements before they are accumulated into the costmap, reducing accepted hole, spike, and RMSE rates under severe glare.

## B. Reliability-Target and Reference-Construction Ablation

Reliability-map quality is evaluated using AUPRC and F1 because glare-corrupted pixels are sparse and class imbalance makes raw accuracy misleading. Table IV further evaluates whether the final reliability estimator depends on the soft target, the temporal depth-difference input, and the pose-aligned multi-view reference-depth construction. Compared with binary supervision, the soft target avoids unstable hard labels near glare boundaries. The temporal term improves detection of view-dependent depth failures, while multi-view reference construction reduces the chance that persistent glare artifacts are treated as valid supervision. The default configuration therefore uses soft reliability supervision, RGB-D plus temporal depth difference, and pose-aligned multi-view reference construction in all main experiments.

**TABLE IV**

Reliability-Target And Reference-Construction Ablation

| **Target / input / reference setting** | **AUPRC ↑** | **F1 ↑** | **L2 FOR ↓** | **L2 FSR ↑** | **Stops/10 m ↓** | **Success (%) ↑** | **Observation** |
|---|---|---|---|---|---|---|---|
| Binary target + Δdepth + multi-view ref. | 0.81 | 0.78 | 0.089 | 0.843 | 1.38 | 83.0 | Hard 0/1 supervision unstable near glare boundaries |
| Soft target + RGB-D only + multi-view ref. | 0.84 | 0.82 | 0.074 | 0.862 | 1.15 | 86.8 | Temporal cue removed; view-dependent artifacts harder to detect |
| Soft target + Δdepth + temporal-filter ref. only | 0.73 | 0.70 | 0.121 | 0.806 | 2.08 | 72.5 | Persistent glare leaks into reference; circular supervision risk |
| Soft target + Δdepth + multi-view ref. (default) | 0.89 | 0.87 | 0.056 | 0.897 | 0.85 | 91.4 | Best setting used in all main experiments |

*Note: AUPRC and F1 are computed by thresholding reliability only for metric evaluation. Training uses the stated target formulation. Downstream L2 metrics use the same costmap configuration as Table V. F1 is reported using the validation-selected threshold.*

## C. Costmap Correctness and Navigation Performance

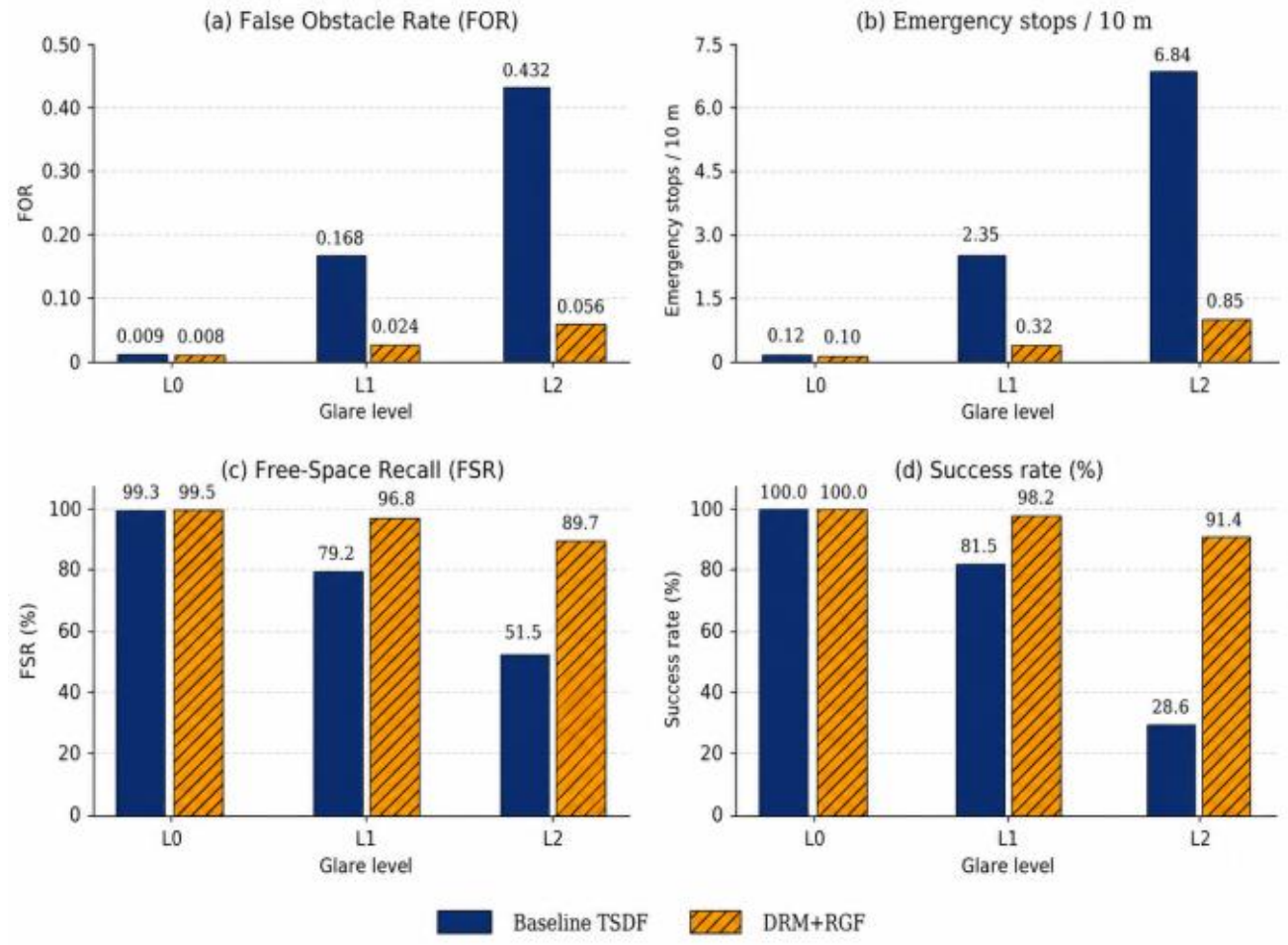

**Fig. 3.** *Quantitative results under three glare severities. DRM+RGF reduces false obstacle rate and emergency stops while improving free-space recall and success rate across L0–L2 conditions.*

**TABLE V**

Costmap Correctness And Navigation Performance

| Glare level | Method | FOR ↓ | FSR ↑ | Emerg. stops/10 m ↓ | Success (%) ↑ |
|---|---|---|---|---|---|
| L0 | Baseline TSDF | 0.009 | 0.993 | 0.12 | 100.0 |
| L0 | Validity/range gating | 0.008 | 0.993 | 0.12 | 100.0 |
| L0 | Spatial median | 0.008 | 0.990 | 0.11 | 100.0 |
| L0 | Temporal rejection | 0.007 | 0.991 | 0.10 | 100.0 |
| L0 | DRM+RGF | 0.008 | 0.995 | 0.10 | 100.0 |
| L1 | Baseline TSDF | 0.168 | 0.792 | 2.35 | 81.5 |
| L1 | Validity/range gating | 0.125 | 0.840 | 1.80 | 88.0 |
| L1 | Spatial median | 0.095 | 0.890 | 1.20 | 92.5 |
| L1 | Temporal rejection | 0.080 | 0.910 | 0.95 | 95.0 |
| L1 | DRM+RGF | 0.024 | 0.968 | 0.32 | 98.2 |
| L2 | Baseline TSDF | 0.432 | 0.515 | 6.84 | 28.6 |
| L2 | Validity/range gating | 0.350 | 0.620 | 5.10 | 45.5 |
| L2 | Spatial median | 0.310 | 0.650 | 4.50 | 52.0 |
| L2 | Temporal rejection | 0.245 | 0.710 | 3.20 | 68.5 |
| L2 | DRM+RGF | 0.056 | 0.897 | 0.85 | 91.4 |

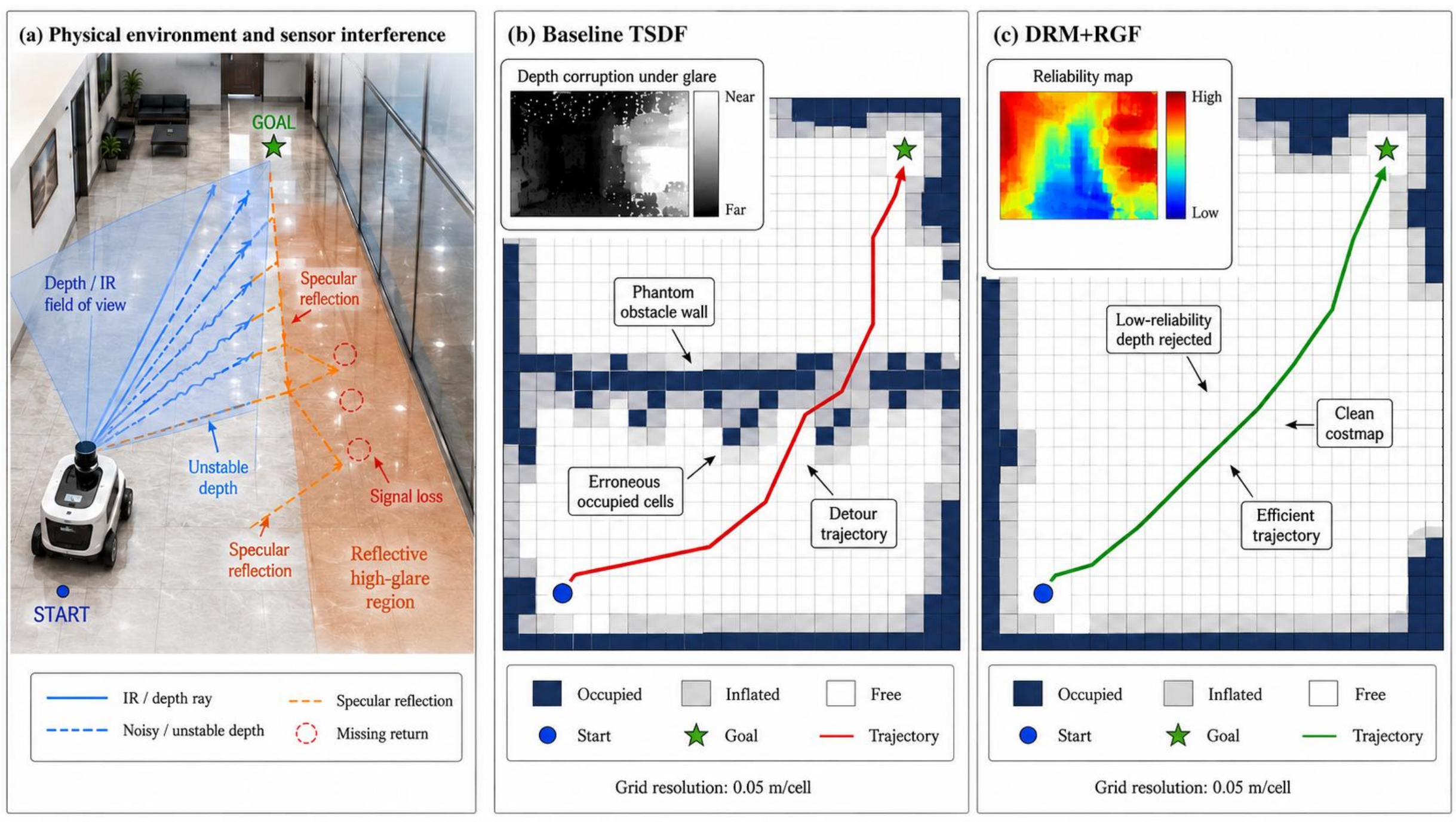

**Fig. 4.** *Trajectory detour analysis under severe glare. Baseline fusion forms glare-induced phantom obstacles that trigger detours, whereas DRM+RGF suppresses unreliable measurements and preserves a clean costmap with an efficient trajectory.*

## D. Fusion-Variant Ablation

**TABLE VI**

Fusion-Variant Ablation Under L2 Glare

| Fusion variant | FOR ↓ | FSR ↑ | Emerg. stops/10 m ↓ | Success (%) ↑ | Main observation |
|---|---|---|---|---|---|
| Baseline TSDF | 0.432 | 0.515 | 6.84 | 28.6 | Phantom wall accumulation |
| Weighted-only | 0.185 | 0.724 | 3.15 | 65.2 | Artifacts attenuated but still persistent |
| Gate-only | 0.048 | 0.785 | 1.10 | 85.5 | Very low FOR but conservative geometry |
| Weighted+Gate | 0.056 | 0.897 | 0.85 | 91.4 | Best balance of safety and free space |

The ablation in Table VI clarifies why the default method uses both weighting and gating. Weighted-only fusion reduces error but still allows persistent low-weight artifacts to accumulate. Gate-only fusion suppresses false positives but removes useful edge geometry, lowering free-space recall. Weighted+Gate provides the best overall navigation success while maintaining low FOR.

## E. Parameter Sensitivity Analysis

Table VII evaluates the temporal forgetting factor λ and reliability threshold $\tau_R$. The default values $\lambda = 0.85$ and $\tau_R = 0.3$ lie in the most stable operating region. Lower λ reacts quickly but causes map flicker and planner restarts; higher λ retains corrupted evidence too long. A lower $\tau_R$ admits more unreliable measurements; a higher $\tau_R$ blocks more true edge geometry.

**TABLE VII**

Parameter Sensitivity Under L2 Glare

| λ | $\tau_R$ | FOR ↓ | FSR ↑ | Emerg. stops/10 m ↓ | Success (%) ↑ | Behavior |
|---|---|---|---|---|---|---|
| 0.75 | 0.3 | 0.082 | 0.915 | 1.85 | 75.5 | Reactive but flickering map |
| 0.95 | 0.3 | 0.145 | 0.810 | 2.60 | 62.0 | Over-persistent occupancy |
| 0.85 | 0.2 | 0.110 | 0.920 | 1.95 | 80.0 | Gate too permissive |
| 0.85 | 0.4 | 0.035 | 0.765 | 1.15 | 82.5 | Gate too conservative |
| 0.85 | 0.3 | 0.056 | 0.897 | 0.85 | 91.4 | Stable operating point |

## F. Cross-Condition Evaluation

To avoid evaluating only one reflective corridor, Table VIII reports held-out test performance across three glare-dominant scene categories. DRM+RGF maintains low FOR and high success across reflective floors, glass boundaries, and glossy tile with natural light, indicating that the model captures active-stereo failure signatures rather than memorizing one visual background.

**TABLE VIII**

Cross-Condition Evaluation Across Glare Scenarios

| Scene | Optical condition | Method | FOR ↓ | FSR ↑ | Stops/10 m ↓ | Success (%) ↑ |
|---|---|---|---|---|---|---|
| Reflective floor corridor | Polished epoxy, overhead LED | Baseline TSDF | 0.380 | 0.610 | 5.50 | 35.0 |
| Reflective floor corridor | Polished epoxy, overhead LED | DRM+RGF | 0.045 | 0.920 | 0.70 | 94.5 |
| Glass wall / automatic door | Transparent glass, side diffuse light | Baseline TSDF | 0.295 | 0.585 | 4.80 | 42.5 |
| Glass wall / automatic door | Transparent glass, side diffuse light | DRM+RGF | 0.068 | 0.865 | 1.10 | 88.0 |
| Glossy tile with window light | IR saturation from natural light | Baseline TSDF | 0.492 | 0.415 | 8.20 | 15.5 |
| Glossy tile with window light | IR saturation from natural light | DRM+RGF | 0.075 | 0.812 | 1.45 | 85.5 |

## G. Paired Completed-Trial Navigation Comparison

Path-length and time-to-goal metrics can be biased when a baseline fails many trials. Under L2, baseline TSDF has a global success rate of only 28.6%, so averages over its successful trials alone may represent unusually favorable cases. The evaluation therefore reports paired completed trials: only start–goal pairs completed by both methods are compared.

**TABLE IX**

Paired Completed-Trial Comparison Under L2 Glare

| Metric | Baseline TSDF | DRM+RGF | Relative improvement / Interpretation |
|---|---|---|---|
| Paired PLR | 1.92 ± 0.45 | 1.06 ± 0.04 | Path redundancy reduced |
| Paired time-to-goal | 85.3 s ± 18.2 s | 48.5 s ± 3.1 s | 43% shorter |
| Paired detour rate | 68% | 8% | Lower replanning frequency |
| Global success rate | 28.6% | 91.4% | Full-sample reference |

*Note: Paired metrics use only the intersection of trials completed by both methods; global success rate is retained as the full-sample safety outcome.*

## H. Runtime and Embedded Deployment

The theoretical FLOP count is complemented by device-level latency. DRM-Net was exported through ONNX and executed as a TensorRT FP16 engine on the Jetson Orin Nano in 15 W mode with a fixed 320 × 240 input. Averaged over 1000 frames, DRM inference required 6.5 ms, and the total pipeline remained faster than the D435 native 30 FPS update rate.

**TABLE X**

Runtime And Computational Overhead

| Method | DRM latency (ms) | Costmap update (ms) | Total (ms) | Throughput (FPS) | Relative overhead |
|---|---|---|---|---|---|
| Baseline TSDF | — | 10.0 | 10.0 | 100.0 | — |
| DRM+RGF | 6.5 | 10.0 | 16.5 | 60.6 | 65.0% |

*Note: GFLOPs in Table I are theoretical; the latency values in this table are measured on the embedded robotic platform.*

# VI. Discussion

## A. Why Reliability-Guided Fusion Instead of Full Depth Completion?

Depth-completion networks can produce visually dense reconstructions, but their objective differs from safe local navigation. In reflective scenes, a dense model may hallucinate plausible but incorrect geometry outside its training distribution. DRM+RGF instead quantifies when the depth sensor should not be trusted and prevents that evidence from entering the map. This design is lighter than full completion and directly aligned with the safety objective of conservative obstacle insertion and free-space preservation [11], [12], [13], [14], [15].

## B. Circular Supervision and Reference-Depth Validity

The primary risk of self-supervised reference-depth construction is circular supervision: persistent glare artifacts may survive temporal filtering and become false targets. The multi-view reprojection check reduces this risk by requiring consistency under camera motion. In addition, geometry-derived reference costmaps remain independent of the sensor stream and are used for downstream FOR/FSR evaluation.

## C. Failure Cases and Degraded Sensing Mode

DRM+RGF may under-insert obstacles when most of the view is consistently flagged as unreliable, such as near large mirrors, wet reflective floors, or glass façades. In this degraded sensing mode, obstacle suppression reduces false positives but also reduces geometric observability. A practical safeguard is to trigger a slow-down and re-observation policy when the low-reliability pixel ratio exceeds a safety threshold, allowing the robot to gather multi-view evidence before continuing.

## D. Extensions to Multi-Sensor Robustness

Reliability modeling is broadly applicable to sensor fusion under adverse conditions. Future work will incorporate mirror/specular segmentation, uncertainty-aware multi-view fusion, and cross-sensor reliability weighting across RGB-D, LiDAR, and event or polarization sensors [18], [23], [24].

# VII. Conclusion

This paper presented a reliability-guided depth fusion framework for suppressing specular-glare-induced corruption before it propagates into TSDF-based mapping and navigation costmaps. The method uses a weighted update with a minimum reliability gate, adopts soft reliability targets, and incorporates pose-aligned multi-view reference-depth construction to reduce circular-supervision risk. Experiments on a real RealSense D435/Jetson Orin Nano robotic platform show improved sensor-level depth integrity, costmap correctness, navigation safety, cross-condition robustness, and embedded real-time performance.

Under severe glare, DRM+RGF reduces FOR from 0.432 to 0.056 and increases FSR from 0.515 to 0.897. These gains translate into success-rate improvement from 28.6% to 91.4% and a reduction in emergency stops from 6.84 to 0.85 per 10 m. Paired completed-trial analysis further shows that, even among trials completed by both methods, DRM+RGF reduces detours and time-to-goal. These results indicate that treating glare as a reliability-estimation problem offers a practical and lightweight path toward safer indoor robot navigation.